\documentclass{ametsocV6.1}
\usepackage{graphicx}
\usepackage{ifpdf}
\usepackage{cite}
\usepackage{amssymb}
\usepackage{bbm}
\usepackage{siunitx}
\usepackage{multirow}
\usepackage{amsmath}
\usepackage{amssymb}
\usepackage[colorlinks=true,linkcolor=red, citecolor=blue, urlcolor=blue,pdfauthor=author]{hyperref}
\usepackage{algorithmic}
\usepackage{array}

\usepackage{stfloats}
\usepackage{url}
\usepackage{lineno}

\nolinenumbers

\title{Global Precipitation Nowcasting of Integrated Multi-satellitE Retrievals for GPM: A U-Net Convolutional LSTM Architecture}

\authors{Reyhaneh~Rahimi \aff{a}, Praveen Ravirathinam \aff{b}, Ardeshir~Ebtehaj \aff{a}, Ali Behrangi \aff{c}, Jackson Tan \aff{d}, and Vipin Kumar \aff{b}\correspondingauthor{Ardeshir Ebtehaj, email@ebtehaj@umn.edu}}

\affiliation{\aff{a}{Department of Civil Environmental and Geo-Engineering and the Saint Anthony Falls Laboratory, University of Minnesota}\\
\aff{b}{Department of Computer Science and Engineering, University of Minnesota} \\
\aff{c}{Department of Hydrology and Atmospheric Sciences, The University of Arizona}\\ 
\aff{d}{Earth Sciences Division, NASA's Goddard Space Flight}}

\abstract{This paper presents a deep learning architecture for 30~\si{min} global precipitation nowcasts with a 4-hour lead time. The architecture follows a U-Net structure with convolutional long short-term memory (ConvLSTM) cells along with a unique recursive ConvLSTM-based skip connection technique to maintain the sharpness of precipitation fields and reduce information loss due to the pooling operation of the U-Net. The training uses data from the Integrated MultisatellitE Retrievals for GPM (IMERG) and a few key drivers of precipitation from the Global Forecast System (GFS). The impacts of different training loss functions, including the mean-squared error (regression) and the focal-loss (classification), on the quality of precipitation nowcasts are studied. The results indicate that the regression network performs well in capturing light precipitation ($<$1.6~\si{mm.hr^{-1}}) while the classification network can outperform the regression counterpart for nowcasting of high-intensity precipitation ($>$8~\si{mm.hr^{-1}}), in terms of the critical success index (CSI). It is uncovered that the inclusion of the forecast variables can improve precipitation nowcasting, especially at longer lead times in both networks. Taking IMERG as a relative reference, a multi-scale analysis, in terms of fractions skill score (FSS), shows that the nowcasting machine remains skillful for precipitation rate above 1~\si{mm.hr^{-1}} at the resolution of 10~\si{km} compared to 50~\si{km} for GFS. For precipitation rates greater than 4~\si{mm.hr^{-1}}, only the classification network remains FSS-skillful on scales greater than 50~\si{km} within a 2-hour lead time.}

\begin{document}
\maketitle
\statement 
This study presents a deep neural network architecture for global precipitation nowcasting with a 4-hour lead time, using sequences of past satellite precipitation data and simulations from a numerical weather prediction model. The results show that the nowcasting machine can improve short-term predictions of high-intensity precipitation on a global scale. The outcomes of the research will enable us to expand our understanding of how modern deep learning can improve the predictability of extreme weather and benefit flood early warning systems for saving lives and properties.  

\section{Introduction}
Precipitation nowcasting, short-term forecasts with a lead time of fewer than 6 hours, is important for weather-dependent decision-making \citep{sun2014use,shi2015convolutional,otsuka2019gsmap}, especially in relationship with early flood warning systems \citep{imhoff2022large}. Modern Numerical Weather Prediction (NWP) models, such as the NOAA's Global Forecast System \citep[GFS,][]{NCEP2021a} provide short-term hourly predictions of near-surface global precipitation at $0.25^\circ$ grid size \citep{toth1997ensemble, sun2014use}. However, NWPs' precipitation nowcasting often faces significant uncertainties due to errors arising in initialization \citep{hwang2015improved}, data assimilation \citep{barker2004three}, and sub-grid scale parameterization of convective processes \citep{zhang2021operational,sun2005convective}. 

Initialization and data assimilation of NWPs often involve a spin-up period of 3--6 hours \citep{buehner2020non, lin2020improving, lin2005precipitation} within which the forecast accuracy is suboptimal \citep{kendon2021challenges}. Moreover, precipitation is not a fundamental state variable (e.g., velocity, temperature, and moisture) in NWPs \citep{lin2015dynamical}. Consequently, variational data assimilation frameworks rely on linear tangent models and their adjoint operators. Since precipitation processes have a flow-dependent non-linear relationship with the basic state variables, even with a carefully designed adjoint operator, the nowcasting uncertainties can remain large \citep{errico2007issues, zhang2013assimilation}. At the same time, the global precipitation forecasts are not currently convective permitting \citep{prein2015review} and rely on sub-grid scale parameterizations \citep{hohenegger2009soil,kendon2012realism}. The parameterization uncertainties often lead to position errors in the prediction of the extent, intensity, and phase of precipitation events, especially for summertime mesoscale convective systems \citep{ban2014evaluation,prein2015review,lin2015dynamical} and short-lived snowstorms \citep{liu2011high}. 

Experimental evidence has suggested that statistical post-processing and machine learning models can temporally extrapolate the observed precipitation fields by ground-based radars \citep{bowler2004development, ayzel2019optical} and spaceborne platforms \citep{otsuka2016nowcasting, kumar2020convcast} with a higher degree of accuracy than NWPs, over limited forecast lead times \citep{sonderby2020metnet}. Data-driven precipitation nowcasting can be grouped into two classes: (i) advective-based approaches  \citep{ayzel2019optical} that often rely on classic regression and optimal filtering \citep{germann2002scale,reyniers2008quantitative} and (ii) modern machine learning (ML) and artificial intelligence (AI) frameworks \citep{ravuri2021skilful,shi2015convolutional}.   

The advective approaches \citep{bowler2004development,pulkkinen2019pysteps} attempt to track the near-surface wind velocity fields and precipitating cells \citep{austin1974use} using optical-flow methods \citep{bowler2004development, bowler2006steps}. Utilizing this approach, different operational nowcasting models have been developed including the {\it Pysteps} \citep{pulkkinen2019pysteps} and the {\it rainymotion} \citep{ayzel2019optical}. These models can only represent a shifted version of previous observations and thus fail to simulate the dynamic evolution of precipitation fields \citep{agrawal2019machine}. However, for example, the results by \citet{lin2005precipitation} showed that the probability of precipitation detection can be around 85 (50)\% for the advection-based nowcasting, compared with 50 (60)\% for the Weather Research and Forecasting \citep[WRF,][]{powers2017weather} model for a 1(6)-hour lead time.

The second class largely relies on deep learning neural networks \citep[DNN,][]{lecun2015deep}, primarily tailored for solving computer vision problems in image classification \citep{rawat2017deep}, video frame prediction \citep{mnih2015human,yang2019video}, and semantic segmentation \citep{mccormac2017semanticfusion}. In summary, a sequence of observed precipitation fields serves as inputs/outputs of a DNN to train it as a nowcasting machine \citep{shi2015convolutional}. Research suggests that these machines can provide more accurate predictions of convective heavy precipitation compared to the advective counterparts \citep{shi2015convolutional, agrawal2019machine, ravuri2021skilful, TREBING2021178, ehsani2022nowcasting}, largely due to their ability to learn complex nonlinear processes. For example, \citet{pulkkinen2019pysteps} highlighted that advection models can offer potentially valuable forecasting skills for precipitation rates of 0.1 \si{mm.hr^{-1}} for a duration of up to 3 hours. However, for more intense rates (i.e., 5 \si{mm.hr^{-1}}), useful nowcasts are only attainable for a maximum lead time of 45 minutes.

The mainstream ML models often use the convolutional neural network \citep[CNN,][]{krizhevsky2012imagenet} and the recurrent neural network \citep[RNN,][]{hochreiter1997long}. The most commonly used CNN nowcasting models are based on U-Net \citep{ronneberger2015u} architecture and its variants \citep{han2021convective, ayzel2020rainnet, TREBING2021178}, where spatiotemporal dynamics of precipitation are captured by convolution operations. Recurrent neural networks \citep{hochreiter1997long}, originally developed for modeling texts can capture temporal memory of the underlying processes. A combination of CNN and RNN namely the convolutional long short-term memory (ConvLSTM) architectures \citep{shi2015convolutional} have been deployed to simultaneously learn spatiotemporal dynamics of precipitation fields for providing regional nowcasts with a lead time of 90 minutes using ground-based radar data. Building on ConvLSTM, multiple variants have emerged to process the spatiotemporal rainfall data such as TrajGRU \citep{shi2017deep} and PredRNN \citep{wang2017predrnn}.

More recently, \citet{ravuri2021skilful} introduced a deep generative neural network for probabilistic nowcasting using ground-based radar data over the United Kingdom that outperformed an advective approach \citep{pulkkinen2019pysteps} for lead times beyond 30 minutes. Despite its success, the model exhibited large uncertainties in the nowcasting of extreme precipitation and cloud dissipation at long lead times. To address this limitation, \citet{zhang2023skilful} proposed NowcastNet that informs the deep-learning machine by the continuity equations and extends the protective skills of the approach by \citet{ravuri2021skilful} -- especially for rates above 16 \si{mm.hr^{-1}}. More recently, \citet{jin2023spatiotemporal} used an encoder-decoder architecture built around ConvLSTM layers to highlight the impacts of temperature, humidity, and wind velocity on precipitation nowcasts.

Most of the developed nowcasting models used ground-based regional precipitation radar observations \citep{han2021convective, sun2014use, pan2021improving}. Recently a few studies have employed satellite data based on advective \citep{kotsuki2019global,otsuka2016nowcasting} and ML frameworks \citep{kumar2020convcast,ehsani2022nowcasting}. Using a local ensemble transform Kalman filter, a prototype system was developed by \citet{otsuka2016nowcasting} and tested using the Japan Aerospace Exploration Agency’s Global Satellite Mapping of Precipitation (GSMaP) \citep{kubota2007global}. More recently, the Integrated Multi-satellitE Retrievals for GPM \citep[IMERG,][]{huffman2020integrated} were used over the eastern CONUS for training and testing both a convolutional deep neural network and a recurrent neural network for precipitation nowcasting with 90 minutes lead time \citep{ehsani2022nowcasting}. 

In this paper, we present a Global prEcipitation Nowcasting using intEgrated multi-Satellite retrIevalS for GPM (GENESIS) that can effectively forecast short-term spatiotemporal dynamics of global precipitation. The GENESIS architecture replaces standard convolution layers in U-Net with ConvLSTM layers, to capture the space-time nature of the precipitation nowcasting problem. Unlike previous works, at the bottleneck layer, we forecast the future latent embeddings, through a recursive generative process, which are then passed to a decoder. We extend this generative forecasting technique to the skip connections, by appending the recursively generated embeddings from each encoder level with the embeddings at their respective level in the decoding blocks. This allows for improved capturing of features at each decoder level and leads to less information loss due to the pooling operations. The network is trained using global IMERG observations \citep{huffman2020integrated}, as well as the GFS forecasts \citep[GFS,][]{NCEP2021a}, providing key drives of precipitation in time and space. Beyond the widely used mean-squared error (MSE) loss function that can lead to underestimations of precipitation extremes, we examine the impacts of the focal loss function \citep[FL,][]{lin2017focal}. This loss function enables capturing the shape of precipitation distribution through a multi-class representation of precipitation rates beyond second-order penalization of nowcasting error, promising to improve nowcasting of precipitation extremes \citep{huang2023stamina}.

Section~\ref{sec:2} explains the data, preprocessing operations, and the mutual information between the inputs and outputs of GENESIS. The methodology is explained in Section~\ref{sec:3} and the results and discussion are presented in Section~\ref{sec:4}. Section~\ref{sec:5} concludes and conjures up future research directions.  

\section{Data, Preprocessing, and Input Feature Analysis}\label{sec:2}

\subsection{Data}
As previously noted, the inputs to GENESIS comprise a temporal sequence of IMERG precipitation, along with physically relevant variables from GFS by the National Centers for Environmental Prediction \citep[NCEP,][]{NCEP2021a}. The half-hourly (hourly) IMERG (GFS) training data are collected from April 2020 to March 2023. The data over the first two years are used for training and the rest for testing.

The IMERG integrates passive microwave \citep{kummerow2015evolution} and infrared \citep{hong2004precipitation} precipitation from the GPM constellation and geostationary satellites. The passive microwave precipitation is mostly processed by the Goddard PROFiling algorithm \citep[GPROF,][]{kummerow2015evolution} and is merged with infrared precipitation \citep{hong2004precipitation} to fill the temporal gaps, using a Kalman filtering approach \citep{joyce2011kalman}. The IMERG product provides gridded precipitation rates globally at $0.1^\circ$ every 30 minutes. The product is available at three different time latencies: Early Run, Late Run, and Final Run. For this study, sequences of half-hourly Early Run (V06) are used with a minimum time latency.

The NCEP's GFS constitutes an atmospheric model that is coupled with the Noah land surface model with multi-parameterization options \citep[NoahMP,][]{niu2011community}, the Modular Ocean Model version 6 \citep[MOM6,][]{adcroft2019gfdl}, and the Los Alamos sea ice model \citep[CICE6,][]{hunke2017cice}. The GFS atmospheric model uses the Finite Volume Cubed-Sphere Dynamical model (FV3) developed by the Geophysical Fluid Dynamics Laboratory
\citep[GFDL,][]{putman2007finite}. In the vertical dimension, there are 127 pressure levels with the top layer centered around 80~\si{km} height. The initial conditions are provided through the Global Data Assimilation System \citep[GDAS,][]{abreu2012description}, which ingests various satellite and ground-based observations using a 4-Dimensional system \citep{lorenc2015comparison}. Here we confine our considerations to the hourly forecasts (v16) of 2-m horizontal $(U,\,V)$ wind velocities in \si{m.s^{-1}} and total precipitable water (TPW, \si{kg.m^{-2}}) content at resolution $0.25^\circ$.

These three physical variables are primarily selected for two main reasons while keeping the computational burden of the training process under control. Firstly, wind velocity plays a pivotal role in the atmospheric transport of moisture and thus the evolution of precipitation cells \citep{huang2023stamina, jin2023spatiotemporal}. Secondly, forecasts of TPW identify the temporal evolution of moisture-laden convective clouds and thus potentially improve the predictive skill of the nowcasting machine, especially in relationship with extreme precipitation \citep{kim2022linking}.

\subsection{Preprocessing for Training}\label{sec:3-1}

\begin{figure} \centering 
\includegraphics[width=0.5\linewidth]{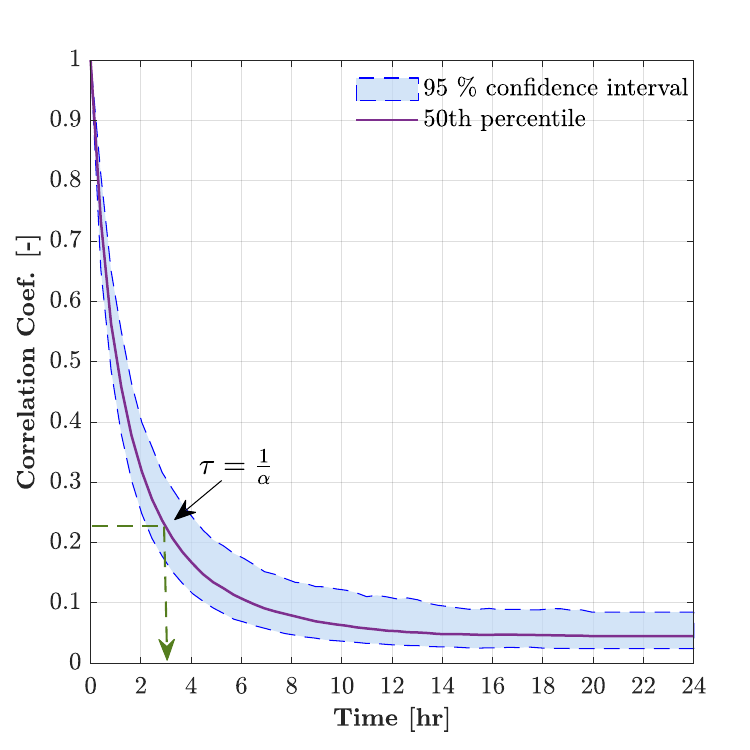}
\caption{Correlation coefficients between IMERG precipitation snapshots at different time lags show an exponential decay rate of $\alpha\approx0.33$ $\text{hr}^{-1}$ and a correlation length $\tau$ of almost 3 hours.}
\label{fig:1}
\end{figure}

The question is how many sequences of IMERG are needed to be fed into the nowcasting machine. The answer to this question may be obtained through auto-correlation analysis. Fig.~\ref{fig:1} shows the median and 95\% confidence bound of the temporal correlation at different time lags over a 24-hour time window, using 1,000 randomly selected IMERG precipitation snapshots.

The analysis shows that the correlation decays almost exponentially in time with a mean rate of $\alpha\approx0.33$~\si{hr^{-1}}, equivalent to a correlation length of $\tau=1/\alpha\approx3$~\si{hr}. As is evident, after 6 hours, the median of the coefficients is around 0.1 and reaches a plateau of 0.05 after approximately 12 hours. Therefore, it appears that beyond 6 to 8 hours, the global precipitation fields have minimal memory of the past, even though a negligible long-range dependency appears to persist. Given this information and the computational complexity of the problem, we confine our considerations to a time window of six hours for nowcasting the next four hours every 30 minutes.  

\begin{figure}[b]
\centering 
\includegraphics[width=0.75\linewidth]{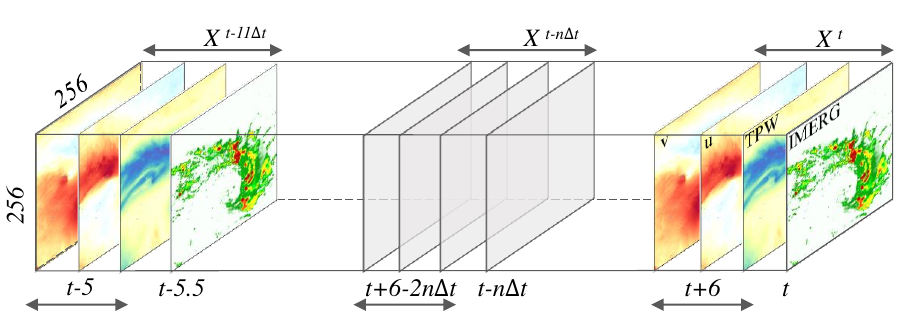}
\caption{Schematic of the input tensors $X^{t-n\Delta t}$ over 6 hours with $n=0,\ldots,11$ -- organized for training the nowcasting machine with $\Delta t = 1/2$~\si{hr}. The tensors contain past IMERG precipitation as well as past and future GFS data of TPW and near-surface horizontal wind velocities components $U$ and $V$.}
\label{fig:2}
\end{figure}

Mapping the GFS variables to the IMERG grid using the nearest-neighbor interpolation, over each grid of the training data at time $t$ there is $C=4$ number of predictors including IMERG precipitation $P$, TPW, $U$, and $V$ -- stored in the channel dimension of a tensor array $X^t \in \mathbb{R}^{M \times N \times C}$. Unlike IMERG precipitation, GFS forecasts are available on an hourly time scale. Therefore, for twelve past sequences of IMERG, we use six past and future sequences of GFS data and organize them for training (Fig.~\ref{fig:2}) as follows:

\begin{equation}\centering
\begin{split}
    X^t & = \left[P^t|\, TPW^{(t+12\Delta t)}|\, U ^{(t+12\Delta t)} | V^{(t+12\Delta t)}\right] \\
    X^{(t-\Delta t)} & = \left[P^{(t- \Delta t)}|\, TPW^{(t+10\Delta t)}|\, U ^{(t+10\Delta t)} | V^{(t+10\Delta t)}\right] \\
     & \vdots \\
    X^{t-n \Delta t} & = \left[P^{(t-n\Delta t)} | \, TPW ^{(t+6-2n \Delta t)} | U^{(t+6-2n \Delta t)} | \, V^{(t+6-2n \Delta t)}\right],
\end{split}
\end{equation}
where $n = 0,1, \dots,11$ with $\Delta t = \frac{1}{2}$\,\si{hr}. By stacking multiple tensors over 6 hours, we have the entire training block of input predictors as $\mathcal{X} = \left[X^{t} | X^{t- \Delta t} | \dots | X^{t-11 \Delta t}\right] \in \mathbb{R}^{M\times N \times C \times T}$. This input block of data is fed into the network to predict the output tensor $\mathcal{Y} = \left[P^{t+\Delta t} | P^{t+ 2\Delta t} | \dots | P^{t+8\Delta t}\right] \in \mathbb{R}^{M \times N \times T_f}$ that contains the next 4 hours of future IMERG precipitation such that 
 $\mathcal{Y} = f(\theta,\mathcal{X}): \mathbb{R}^{M \times N \times C \times T} \rightarrow \mathbb{R}^{M \times N \times T_f}$, where $\theta$ denotes all the unknown network parameters.   

Here, we choose patches of size $M=N=256$ to make the training process computationally amenable. The outputs are then tiled with 50\% overlap and only $128\times128$ interior pixels are used to reconstruct the entire domain of IMERG. Overall, 100,000 patches of tensor blocks were randomly sampled over the spatial domain of the problem from April 2020 to March 2022. This data set was split into 70 and 30\% for training and validation. For testing 15,000 independent random samples were used from April 2022 to March 2023.

\begin{figure*}[t!]
\centering
\includegraphics[width=0.8\linewidth,]{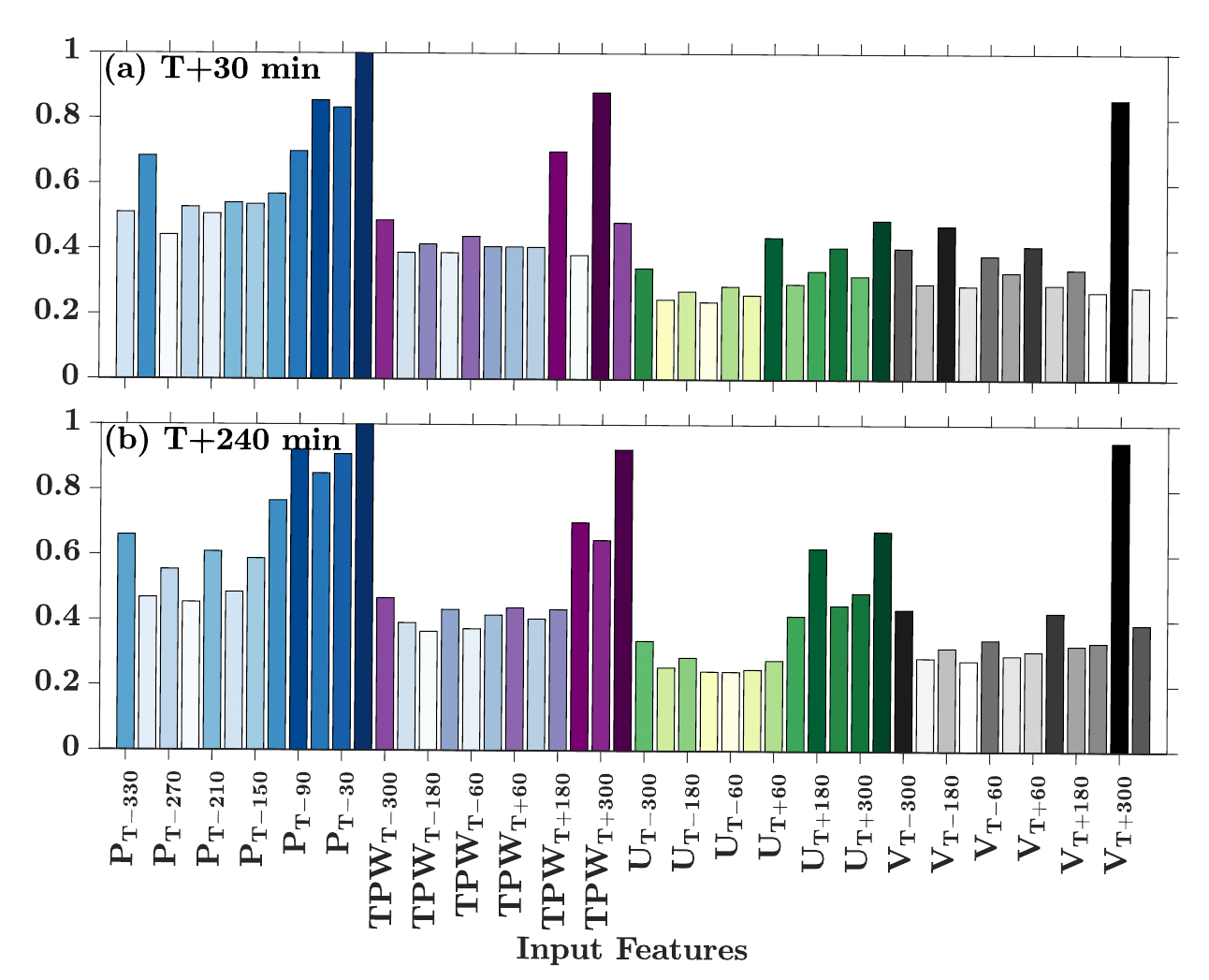}
\caption{Minimum redundancy and maximum relevance (MRMR) scores of inputs for precipitation nowcasting at (a) first, and  (b) last lead time.}
\label{fig:3}
\end{figure*}

\subsection{Input Feature Analysis}\label{sec:3-3}

Before delving into the design of the nowcasting machine, a key question emerges: Amongst the used predictors, what are the most important ones concerning the short-term evolution of precipitation fields? Here, to shed light on this question, the maximum relevance and minimum redundancy method \citep[MRMR,][]{peng2005feature} is employed. This method identifies the most informative and the least redundant features in a given set of predictors, irrespective of the underlying ML algorithm. Specifically, it finds an optimal set $S$ of features that maximizes $A_s = 1/|S|\sum_{x\in S}I(x,y)$, the relevance of features in $S$ with respect to a response variable $y$, and minimizes $B_s = 1/|S|^2\sum_{x,z\in S}I(x,z)$, the redundancy of features in $S$ -- where $I(\cdot,\cdot)$ is the mutual information and $|S|$ denotes the cardinality of $S$.  Those predictors with high MRMR scores exhibit the highest mutual information with the temporal evolution of precipitation fields and the least redundancy in terms of their internal mutual information.

The results of the MRMR analysis at both the initial (a) and final (b) lead times are presented in Fig.~\ref{fig:3}. The figure illustrates that, at the first lead time, on average, previous precipitation fields exhibit a higher mean score (0.64), followed by  TPW (0.48) and horizontal wind velocity (0.32). Within the precipitation sequence, the first 2 hours display a higher MRMR score, and the score gradually decreases as we move backward in time. For the TPW, the future sequences $T+150$ -- $T+300$ \si{min} demonstrate the highest scores, highlighting their predictive impacts for longer lead times. Moreover, the scores for $U$ are larger than $V$ and increase for longer lead times, as the global storm morphology and evolution are chiefly governed by the prevailing easterlies and westerlies.  

Examining the mutual information between inputs and precipitation fields at a 4-hour lead time (Fig.~\ref{fig:3}-b) reveals that still, previous sequences of precipitation are more important than the used physical variables. However, the mean MRMR score for physical variables shows an increase of 6, 12, and 4\% for TPW, $U$, and $V$, respectively, compared to those for the 30-minute lead time. This further highlights the role of horizontal velocity forecasts in the nowcasting of precipitation for longer lead times.

\section{Methodology}\label{sec:3}

\subsection{Network Architecture}

The used architecture fuses a U-Net \citep{ronneberger2015u} and a Convolutional LSTM \citep[ConvLSTM,][]{shi2015convolutional} neural network. The standard U-Net encodes the semantic information of a single image into a low-dimensional latent representation, which is then decoded to produce the segmentation results along with the help of skip connections. These connections help to recover fine-level details lost in the encoding stage and converge much faster to the optimal estimate of the network parameters while alleviating the vanishing gradient problem \citep{lecun2015deep}. However, the U-Net architecture is primarily designed to learn spatial features of static fields and has no specific mechanisms for capturing temporal dependencies in data sequences \citep{wang2022predrnn}. 

For time series modeling with temporal dependencies, LSTM networks extend the predictive skills of classic RNNs \citep{hochreiter1997long}. The LSTM memory cells enable the retention of important dependencies and propagate the error information over a sufficiently long window of time for updating its parameters without the vanishing and/or exploding gradients problems \citep{hochreiter1997long, graves2013generating}. However, LSTMs work mainly with time series data, which are not spatial in nature. To overcome this issue, the Convolutional LSTM (ConvLSTM) cell was introduced \citep{shi2015convolutional} for time series analysis of spatial domains. In the context of precipitation nowcasting, it is important to note that while the dynamics of the involved atmospheric variables may not necessarily exhibit long-range temporal dependencies, it is imperative to deploy modern LSTM networks to account not only for potential long-range dependencies but also the existing short-range correlations.

\begin{figure*}
\includegraphics[width=1\linewidth]{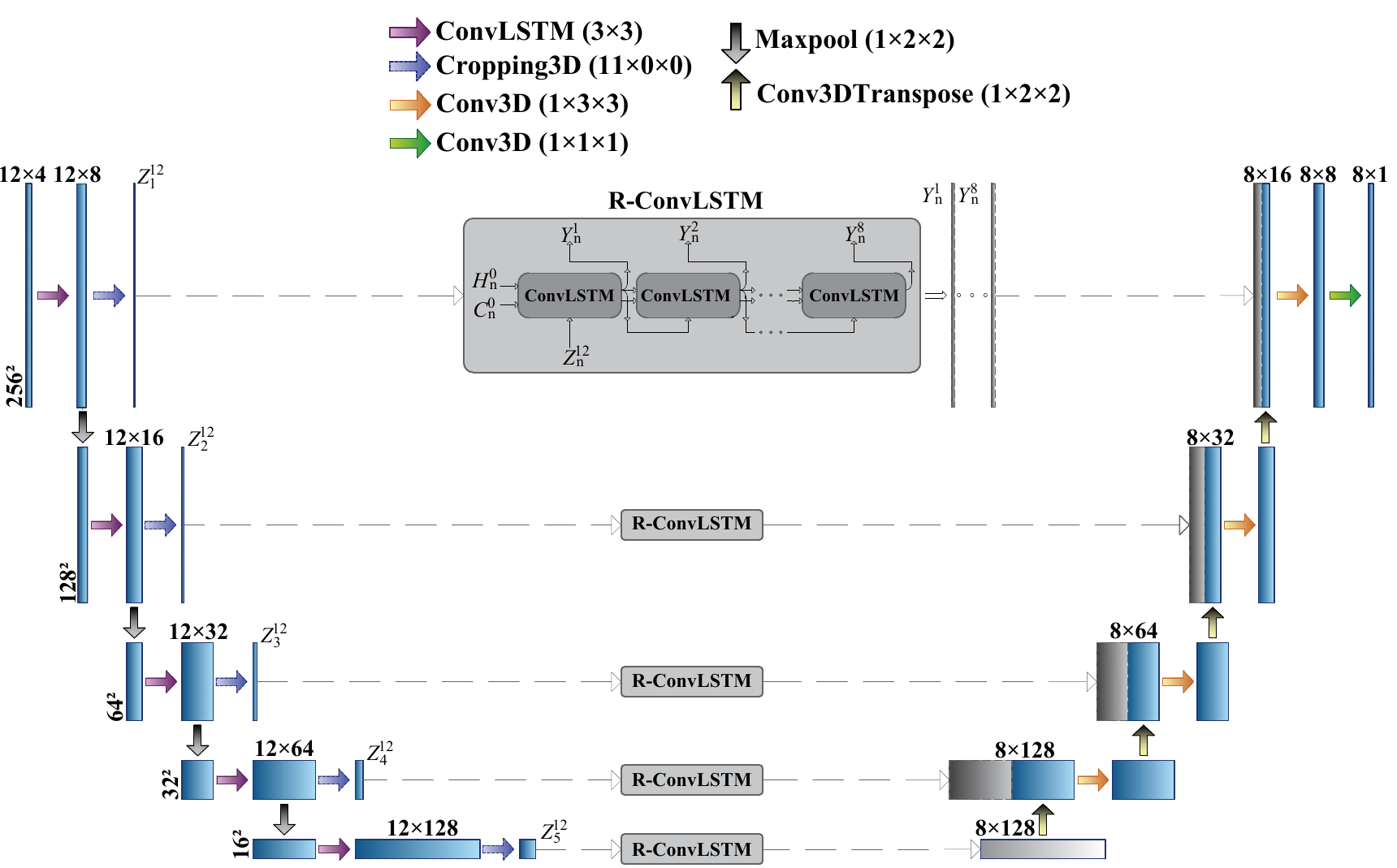}
\caption{Schematics of the GENESIS nowcasting machine. An encoder operates over a sequence of input data, at 12 time steps, through a series of convolutional LSTM kernels to extract their low-dimensional spatiotemporal features and then a decoder upsamples the features through a series of deconvolution to reconstruct the evolution of precipitation fields over a 4-hour lead time at 8 time steps.}
\label{fig:4}
\end{figure*}

To harness the combined capabilities of U-Net and ConvLSTM, we propose a new architecture for IMERG precipitation nowcasting (Fig.~\ref{fig:4}). In the proposed architecture, each input tensor $X \in \mathbb{R}^{H\times W\times C}$ of a sequence $\mathcal{X} = \left[X^1, X^2, \dots, X^{T}\right]$, with $T=1,\ldots,12$ is passed through a ConvLSTM operator to encode the spatiotemporal information of each timestep into a latent representation $Z^T_n \in \mathbb{R}^{H^{\prime} \times W^{\prime} \times C^{\prime}}$ at $n=1,\ldots,5$ encoder blocks, followed by max-pooling, batch normalization, and 15\% dropout layers.

In the last encoding block, the nowcasting machine passes the latent embeddings of the preceding ConvLSTM $Z_{n=5}^{12}$ through a Recursive ConvLSTM (R-ConvLSTM) operator that forecasts the evolution of the embeddings over the next 4-hour lead time at eight timesteps. As sketched in Fig.~\ref{fig:4}, this recursive generative process is also repeated for skip connections at multiple encoding blocks which allows for improved capturing of features at each decoder level and leads to less information loss due to the pooling operations.

The Recursive ConvLSTM uses $Z_n^{12}$ as input and considers zero values for the initial hidden $H_n^0$ and cell $C_n^0$ states to compute the output states $Y_n = [Y_n^{1}, \dots, Y_n^{T}]$ for $T=1,\ldots,8$ time steps, recursively. These predicted states are then concatenated with the outputs of a transposed 3D convolution operator (Conv3DTranspose) at each encoder block and passed through a 3D temporal convolution (Conv3D), batch normalization, and 15\% dropout layers.

Note, to ensure that the context of the last encoder latent embedding (i.e. $Z_{n}^{12}$) is not lost while the recursive generation of future latent spaces is happening, we add this final context to each input of the recursive generative process. This technique helps retain the information across all the timestamps of forecast and has been shown to lead to better temporal generation \citep{SLTLAE_CL_paper}. Specifically, at each step of future latent embedding generation, three outputs are given: output features, cell states, and hidden states. The context tensor, $Z_{n}^{12}$, is then added to the output features, to create the joint context. This joint context is passed as input to the next step of the recursive ConvLSTM, with the new cell and hidden states. This process is repeated until the eight steps of future latent embeddings have been created. This creation of the joint context helps preserve information from the encoder over multiple steps of unrolling.

\subsection{Loss Functions}\label{sec:3-2}

In the proposed architecture, the choice of the loss function can impact the accuracy of precipitation nowcasting. Different loss functions can be tested for this purpose including the mean squared or absolute error losses, Wasserstein distance \citep{villani2009optimal}, and those loss functions that are often used for classification problems such as the cross entropy \citep{mackay2003information} and FL functions \citep{lin2017focal}. 

Commonly used loss functions such as the MSE cannot capture the non-Gaussian probability distribution of the precipitation fields \citep{guilloteau2023constraining, ehsani2022nowcasting}. Another approach is to discretize the input-output rates of precipitation into a finite number of intervals and recast the nowcasting into a multi-class classification problem. The main advantage of the latter approach is that the network would attempt to implicitly reconstruct the precipitation histogram rather than minimize the second-order statistics of the nowcasting error \citep{pfreundschuh2022improved}.
  
As a commonly used loss function for classification tasks, cross-entropy measures the dissimilarity between two discrete probability distributions. However, its performance degrades markedly, when there is a high imbalance between the number of classes. In this scenario, those classes with a large number of training samples (i.e., low precipitation rates) will dominate the value of the loss function, causing the weights to be updated in a direction that the network becomes less and less skillful in predictions of those classes with lesser training samples (i.e., high precipitation rates). 

To that end, the FL loss function uses a focusing parameter $\gamma$ to avoid saturating the loss with the impacts of easy-to-classify samples and balances the contribution of each class to the total loss by a weighting factor $\alpha_c$ as follows:
\begin{flalign}\label{eq:2}
    \text{FL}(y,\,p)=-\frac{1}{N} \sum_{i=1}^N \sum_{c=1}^C \alpha_c\,y_{i, c}\left(1-p_{i, c}\right)^\gamma \log p_{i, c}
\end{flalign}
where $y_{i,c}$ and $p_{i,c}$ are the reference and predicted probabilities of $c$\textsuperscript{th} class for $i=1,\ldots,N$ training data points. The $\alpha_c$ is proportional to the inverse of the frequency of occurrence of class data points, such that $\sum_c\alpha_c=1$, and $\gamma$ commonly ranges from 2 to 5, with larger values for highly imbalanced training data sets.  

We use both regression MSE and FL loss functions, with $\gamma=2$, for training the nowcasting machine, hereafter often referred to as $\rm GENESIS_{MSE}$ and $\rm GENESIS_{FL}$, respectively. To train $\rm GENESIS_{FL}$, the precipitation rates are discretized into 10 classes in a logarithmic scale ranging from 0.1 to 32 \si{mm.hr^{-1}}. Rates below (above) 0.1 (32) \si{mm.hr^{-1}} are assigned to class 1 (10). Note, that the only difference in the architecture of the network when using MSE and FL loss functions, is the final operation in the last decoder block (Fig.~\ref{fig:4}). This operation employs a layer of $ 1\times1$ convolution followed by a normalization layer and a ReLU activation (Softmax) for the MSE (FL) loss. It is important to emphasize that the $\rm GENESIS_{FL}$ network produces output in the form of posterior probabilities for classes within each bin. These posterior probabilities serve as the foundation for calculating the maximum a-posteriori estimate of the precipitation class through the Softmax function and can be used to detect the exceedance probability of a precipitation threshold \citep{pfreundschuh2022improved}. To obtain the optimal weights of both networks, the Adam optimizer \citep{kingma2014adam} is employed with an initial learning rate $\eta$=1E-3 and the mini-batch size 8. A decay factor of 0.1 is set for the learning rate at every 10 epochs when there is no improvement in the validation loss. To initialize the weights of the nowcasting networks, the {\it Xavier} method \citep{sun2014improving} is used. The network is trained using the TensorFlow-V2 library \citep{abadi2016tensorflow} on an AMD EPYC 7542 with a 32-core central processing unit and 1008 GB of random access memory -- equipped with an NVIDIA A-100 40-GB graphical processing unit, provided by the Minnesota Super Computing Institute (MSI).

\subsection{Nowcasting Quality Metrics} 

To assess the quality of precipitation nowcasts, common categorical evaluation metrics are used including precision $\rm TP/(TP+FP)$, recall $\rm TP/(TP+FN)$, Heidke skill score \citep[HSS,][]{jolliffe2012forecast}, and critical success index ($\rm CSI_{r}$, $TP/(TP+FN+FP)$, \citep{schaefer1990critical}) -- in which the true positive (TP), true negative (TN), false positive (FP), and false negative (FN) rates characterize the confusion matrix. Furthermore, to compare the performance of the models at different spatial scales, fractions skill score FSS=$1-\frac{\sum_{i=1}^M(S_{f}^i -  S_{obs}^i)^2}{\sum_{i=1}^M(S_{f}^i)^2 - \sum_{i=1}^M(S_{obs}^i)^2}$ is used in which $S_{f}$ and $S_{obs}$ are fractional coverage of the nowcasts and observations for $M$ selected windows with a spatial scale of interest \citep{ayzel2020rainnet}. 

To quantify the dissimilarity between the observed and predicted probability distribution of precipitation classes, the Wasserstein distance \citep{villani2009optimal} is employed. This metric is derived from optimal transport theory, which considers the problem of quantifying the cost of morphing one distribution into another non-parametrically. The cost defines a distance metric that not only penalizes the shifts between the mean of the distributions (bias) but also the misshape in terms of all higher-order moments. Smaller values of this metric denote that the two densities are closer in terms of their central locations and shapes. For further information, the readers are referred to \citep{ning2014coping,tamang2020regularized,tamang2021ensemble} in the context of geophysical data assimilation and climate model comparisons \citep{robin2017detecting,vissio2020evaluating}.

The value of precision, recall, $\rm CSI_r$, and FSS can vary from 0 to 1 with a perfect score of 1. The values of HSS vary from $-\infty$ to 1 indicating the worst and the best forecasts while positive values denote better than a random prediction. To compute $\rm CSI_r$ index, the elements of the confusion matrix are determined for a binary representation of the precipitation fields with the rates above a prespecified threshold rate $r$ in \si{mm.hr^{-1}}. To evaluate the performance of the regression network using categorical scores, the predicted rates are converted to the same classes defined for the classification network.

\section{Results and Discussion}\label{sec:4}

It is important to emphasize that the goal of this section is to demonstrate the extent to which we can predict the sub-hourly evolution of the global precipitation field, considering IMERG as a relative and not an absolute reference observation. Therefore, when the nowcasting results are compared with GFS forecasts, we do not aim to quantify the intrinsic errors between the two data sets. However, the presented results are predicated on the presumption that the observational satellite data exhibit less uncertainty than the GFS predictions.

\subsection{Storm-scale Nowcasting}
\begin{figure}[t]\centering
\includegraphics[width=0.8\linewidth]{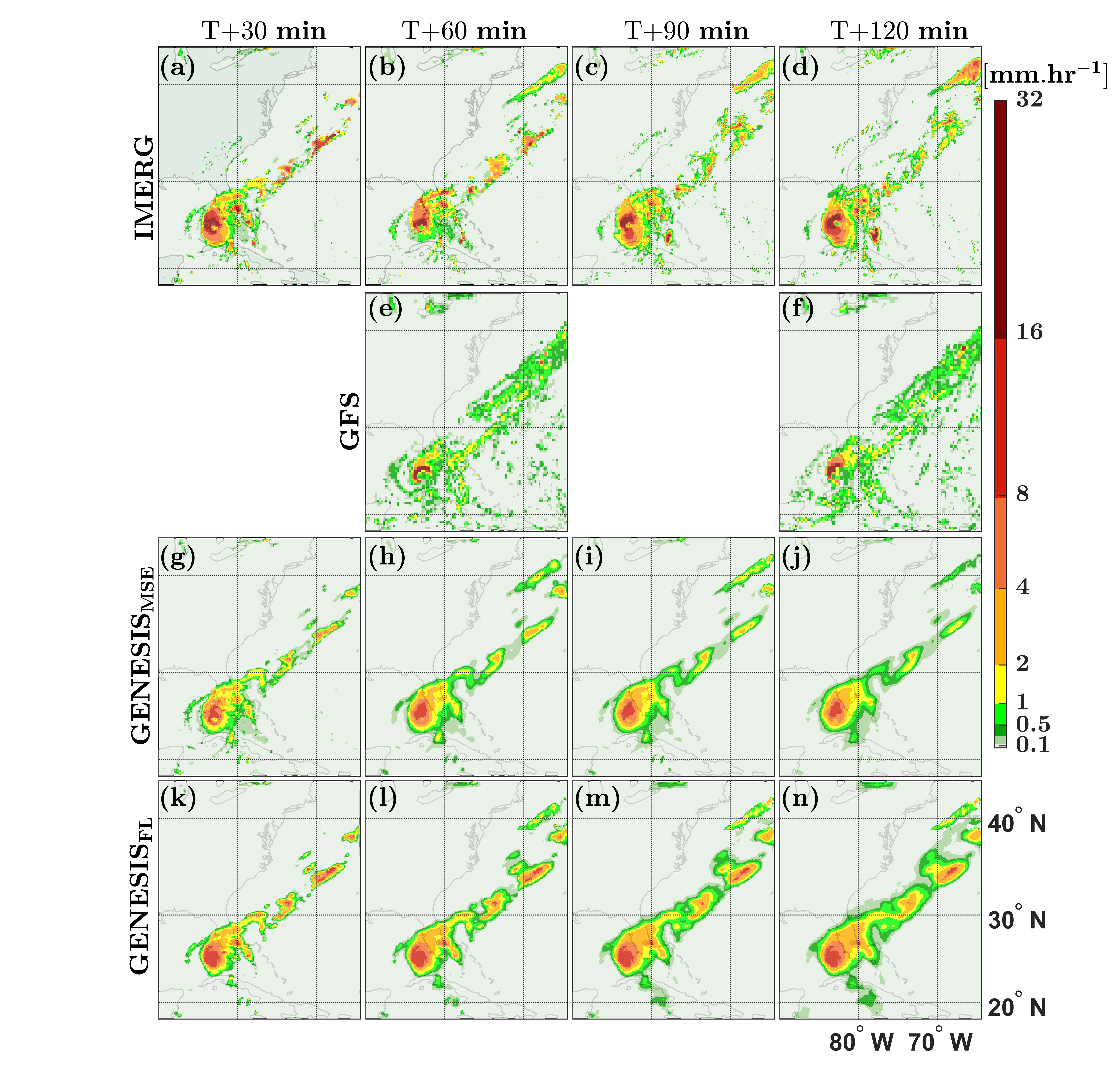}
\caption{Evolution of the Hurricane Ian over 2 hours, with a timestep of 30 min, beginning at 20:00 UTC on September 27, 2022, by IMERG (a--d), GFS (e--f), GENESIS\textsubscript{MSE} (g--j), and GENESIS\textsubscript{FL} (k--n).}
\label{fig:5}
\end{figure}

Figure~\ref{fig:5} shows the nowcasting results of Hurricane Ian initialized at 20:00 UTC on September 27, 2022. Visual inspection indicates that the presented nowcasting machine can predict the storm space-time dynamics. Overall, the GFS predictions are less coherent than IMERG and lack the prediction of some heavy precipitation cells. The extent of the storm in GFS is larger than the IMERG. There are intense precipitation cells in IMERG, organized in the storm band structures, that do not manifest themselves in the GFS forecasts. This is partly expected as the GFS spatial resolution is coarser than IMERG. The difference in capturing high-intense cells is not only apparent over the rainbands around the eyewall but also those extended outer bands, for example near the coasts of South Carolina at the 4\textsuperscript{th} time-step. The nowcasting machine seems to produce precipitation fields that are expectedly consistent with the IMERG observations, even though some underestimations are apparent -- especially over long lead times due to repeated applications of convolution and max-pooling operations.

Comparing the performance of the $\rm GENESIS_{MSE}$ and $\rm GENESIS_{FL}$ reveals that at 30~\si{min} lead time, both models capture some detailed structures of the observed storm, although the $\rm GENESIS_{MSE}$ underestimates the rates, relatively. After an hour, it becomes apparent that the use of the FL function can lead to improved preservation of precipitation extremes when compared to its regression counterpart. It is worth noting that both nowcasts are perceptually closer to the IMERG observations than GFS forecasts. For example, as shown in the dashed red box in Fig.~\ref{fig:5}\,b, an outer rainband manifests itself near the coast of New York and intensifies over longer time steps. This rain band is missed in GFS forecasts; however, is detected and evolved in the nowcasting precipitation products. For longer lead times, the classification network shows a tendency to preserve the precipitation extremes at the expense of overestimating the extent of the storm.     

\begin{table}[h]
\centering

\begin{tabular}{p{1.2cm} c p{1cm} p{1cm} p{1cm} p{1cm} p{1cm} p{1cm} p{1cm} p{1cm}}
\multicolumn{2}{l}{} & \multicolumn{8}{c}{Lead Time} \\ \cline{3-10} \\
Metric & Model & $T+30$ & $+60$ & $+90$ & $+120$ & $+150$ & $+180$ & $+210$ & $+240$   \\ \hline
\multicolumn{1}{c|}{\multirow{3}{*}{$\rm CSI_{1}$}} & \multicolumn{1}{c|}{GFS} &      & 0.25 &      & 0.25 &      & 0.26 &      & 0.28 \\ \cline{3-10} 
\multicolumn{1}{c|}{}                     & \multicolumn{1}{c|}{$\rm MSE$} & 0.75 & 0.69 & 0.63 & 0.58 & 0.47 & 0.41 & 0.37 & 0.35 \\ \cline{3-10} 
\multicolumn{1}{c|}{}                     & \multicolumn{1}{c|}{$\rm FL$}  & 0.71 & 0.65 & 0.60 & 0.56 & 0.47 & 0.43 & 0.39 & 0.39 \\ \cline{3-10} 
\multicolumn{1}{c|}{}                     & \multicolumn{1}{c|}{$\rm BM$}  & 0.67 & 0.47 & 0.42 & 0.38 & 0.35 & 0.33 & 0.32 & 0.31 \\ \hline
\multicolumn{1}{c|}{\multirow{3}{*}{$\rm CSI_{8}$}} & \multicolumn{1}{c|}{GFS} &      & 0.14 &      & 0.24 &      & 0.18 &      & 0.12 \\ \cline{3-10} 
\multicolumn{1}{c|}{}                     & \multicolumn{1}{c|}{$\rm MSE$} & 0.55 & 0.43 & 0.36 & 0.30 & 0.28 & 0.27 & 0.25 & 0.18 \\ \cline{3-10} 
\multicolumn{1}{c|}{}                     & \multicolumn{1}{c|}{$\rm FL$}  & 0.59 & 0.48 & 0.41 & 0.37 & 0.35 & 0.35 & 0.32 & 0.31 \\ \cline{3-10}
\multicolumn{1}{c|}{}                     & \multicolumn{1}{c|}{$\rm BM$}  & 0.50 & 0.39 & 0.32 & 0.20 & 0.18 & 0.18 & 0.15 & 0.14 \\ \hline
\end{tabular}

\caption{Critical success index for GFS, $\rm GENESIS_{MSE}$, and $\rm GENESIS_{FL}$ as well as benchmark persistent (BM) scenario against IMERG observations at threshold 1 and 8 \si{mm.hr^{-1}} for the storm shown in Fig.~\ref{fig:5}.} \vspace{-4mm}
\label{tabel:1} 
\end{table}

\begin{figure}[t]\centering
\includegraphics[width=1\linewidth]{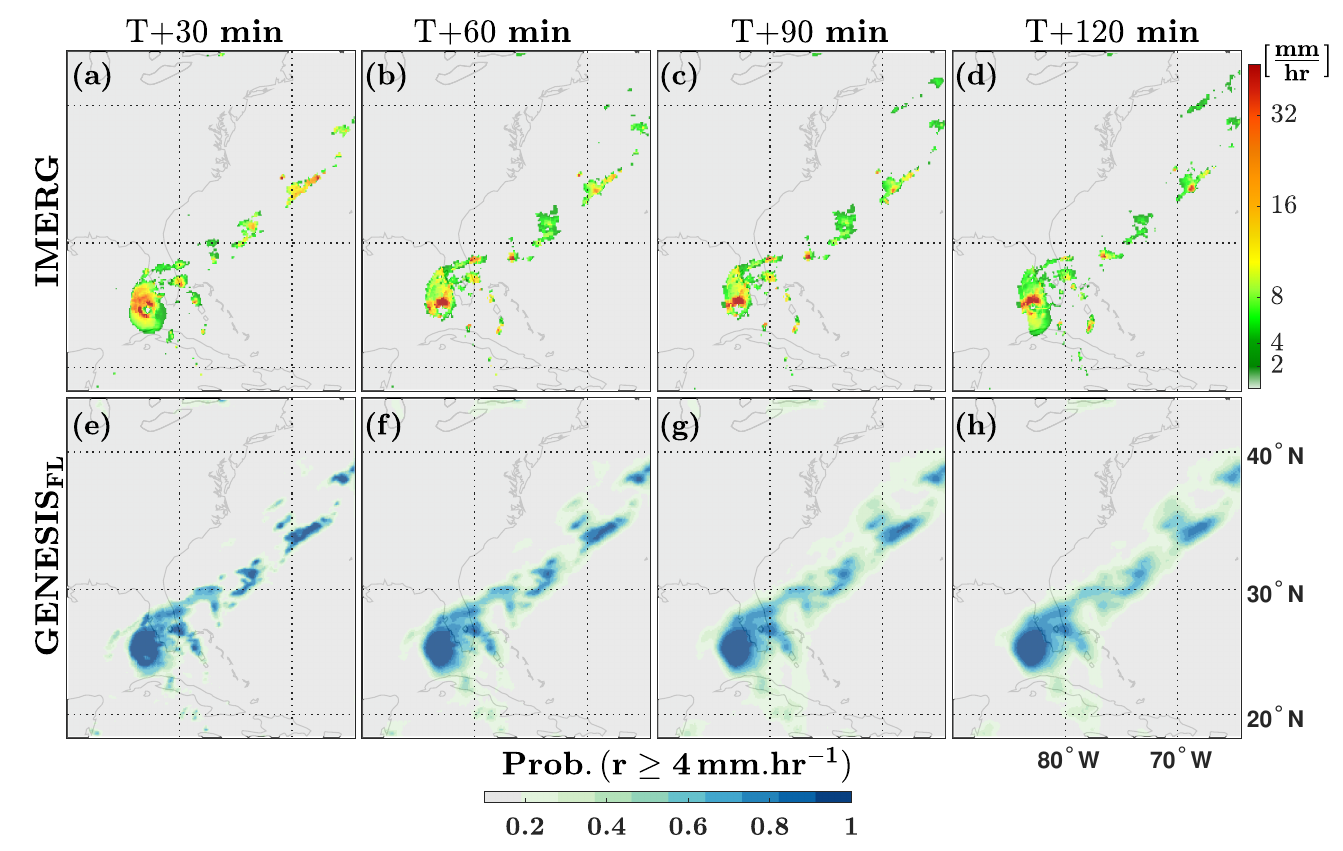}
\caption{Evolution of the Hurricane Ian over 2 hours beginning at 20:00 UTC on September 27, 2022, by IMERG (a--d) masked for the rates above 4 \si{mm.hr^{-1}}, and the corresponding probability of exceedance using $\rm GENESIS_{FL}$ (e--h).}
\label{fig:6}
\end{figure}

Table~\ref{tabel:1} presents $\rm CSI_r$ with two different rates $r=1$ and 8 \si{mm.hr^{-1}} for GFS, GENESIS, and persistence Benchmark (BM) scenario.  The persistence BM scenario assumes that the precipitation field remains time-invariant over the nowcasting time steps, commonly used \citep{ehsani2022nowcasting, ayzel2020rainnet, agrawal2019machine} to test the effectiveness of nowcasting. The results indicate the CSI values for GFS forecasts remain around 0.25 for $r=1$~\si{mm.hr^{-1}} and vary between 0.14 to 0.24 for $r=8$ \si{mm.hr^{-1}}. For the nowcasting machines, the $\rm CSI_1$ score decreases approximately from 0.75 to 0.35 when the lead time grows. The $\rm CSI_1$ score is slightly higher for the $\rm GENESIS_{MSE}$ than the $\rm GENESIS_{FL}$ over lead times less than 180 minutes and then the $\rm GENESIS_{FL}$ gradually overtakes the $\rm GENESIS_{MSE}$ for longer times. Comparing the performance of GENESIS models with the BM scenario shows that both models outperform it by at least 6\% in the first lead time which increases to 42\% at the third time step, in terms of the $\rm CSI_1$ metric. For $r=8$~\si{mm.hr^{-1}}, $\rm GENESIS_{FL}$ provides a superior nowcast quality than $\rm GENESIS_{MSE}$, and the difference tends to grow from 8 to 23\% as the lead time increases from 30 minutes to 2 hours. The CSI value for the BM scenario is lower than the $\rm GENESIS_{MSE}$ models by 10\%  and 28\% at the first and last lead times, which is expected given that the intensity and morphology of high-intense precipitation cells evolve faster in time than those cells with lighter rates. 

An important outcome of $\rm GENESIS_{FL}$ is that it can provide the class probability through the Softmax layer and thus the exceedance probabilities. Figure~\ref{fig:6} a--d shows the storm (first row) over the areas with rates greater than 4~\si{mm.hr^{-1}} and the associated exceedance probabilities (second row). The results show the capability of $\rm GENESIS_{FL}$ in capturing the shape of intense precipitation with high probability. As the lead time increases, the predictions become increasingly blurry, which manifests itself in the probability maps as well.

\subsection{Multi-storm Nowcasting}

\begin{figure*}[b]
\includegraphics[width=1\linewidth,]{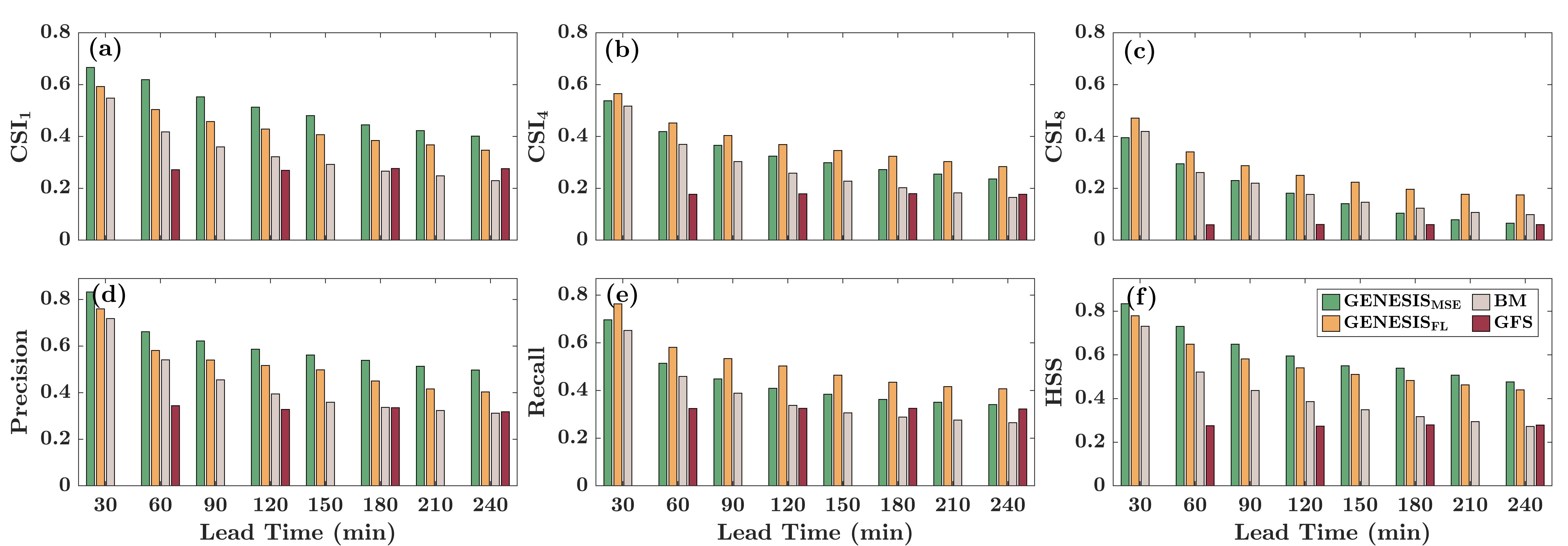}
\caption{Ensemble mean of (a) $\rm CSI_{1}$, (b) $\rm CSI_{4}$, (c) $\rm CSI_{8}$, (d) precision, (e) recall, and (f) HSS for $\rm GENESIS_{MSE}$, $\rm GENESIS_{FL}$, GFS forecasts, and BM scenario using 15,000 independent input samples from the test set.}
\label{fig:7}
\end{figure*}

To provide a more comprehensive evaluation of the nowcasting machines, we randomly selected 15,000 independent samples of input-output tensor blocks, from April 2022 to March 2023 and used them for nowcasting and calculation of the forecast quality metrics. Four different metrics including $\rm CSI_r$ at three threshold values, precision, recall, and HSS are calculated for each sample, and their ensemble mean is presented in Fig.~\ref{fig:7} for different lead times. 

Figure~\ref{fig:7}\,a shows the $\rm CSI_1$ for all three nowcasts and the BM scenario. The GFS performance is fairly stable over different lead times and remains around 0.3. However, $\rm CSI_1$ for regression (classification) network reduces from 0.65 to 0.42 (0.58 to 0.35) when lead time increases from 30 to 240~\si{min}. This difference can be attributed to the imbalanced nature of the data and improved skills of the regression network to represent the bulk of the precipitation distribution. Both GENESIS models outperform the persistence BM scenario, for which $\rm CSI_1=0.54$ at 30~\si{min} and declines rapidly by a factor of 2 at 240~\si{min} lead time.

By increasing the CSI threshold to 4~\si{mm.hr^{-1}} (Fig.~\ref{fig:7}\,b), the performance of GFS forecasts reduces by 30\%, from 0.30 to 0.20, at $T+60$ \si{min} but still remains fairly invariant in time. However, the performance of the regression network decays by a factor of 2, from approximately 0.53 to 0.23, when lead time increases from half an hour to four hours. Meanwhile, the $\rm CSI_4$ in the classification network decays from 0.56 at 30~\si{min} to 0.28 at 240~\si{min}, showing that this network begins to outperform the regression network by capturing higher intensity precipitation rates.

For more intense precipitation (Fig.~\ref{fig:7}\,c), the results indicate that the value of $\rm CSI_{8}$ at the second lead time for GFS, $\rm GENESIS_{MSE}$, $\rm GENESIS_{FL}$, and BM models decay by 65, 27, 25, and 30\% at $T+60$~\si{min} compare to $\rm CSI_4$, respectively. This metric, for the regression network, has its highest value of 0.39 at 30~\si{min} lead time and reduces to 0.1 after three hours. Meanwhile, the classification network still shows some skills and a potential to accurately capture precipitation rates above 8 \si{mm.hr^{-1}}, as $\rm CSI_{8}$ is around 0.47 (0.17) at 30 (240)~\si{min}. It should be noted that the $\rm CSI_{8}$ is low for all predictions and never exceeds 0.4 for the regression network. 

Figure~\ref{fig:7}\,d shows the mean precision metric, which explains the number of truly detected precipitating events divided by all detected precipitating events. The precision value for GFS remains constant in time around 0.35. For the regression (classification) network, it is around 0.83 (0.76) at the first lead time and decays by 30 (32)\% to 0.58 (51) at the fourth time step and then remains relatively constant for longer times. The reason can be attributed to the fact that while the classification network has a higher probability of detection than the regression network, it also exhibits a higher false alarm rate, especially in capturing the more frequent body of precipitation distribution with a rate of less than 1--1.5~\si{mm.hr^{-1}}. The elevated number of false alarm rates leads to a larger denominator in the precision equation, resulting in lower values. As shown, the precision improves over longer lead times relative to the BM scenarios, and the $\rm GENESIS_{MSE}$ consistently outperforms other nowcasts.

The recall metric (Fig.~\ref{fig:7}\,e) shows the ratio of correctly classified sample points to the total observations. The recall value of GFS forecasts is around 0.35 and remains invariant in time. The regression network has a value equal to 0.69 at the first time step which reduces by a factor of 2 at the fifth lead time and remains relatively constant around 0.35. Meanwhile, the metric for the classification network is around 0.75 (0.50) at 30 (150)~\si{min}. The reason can be attributed to the higher skills of this network in detecting high-intense precipitation rates with low false negatives. As for the BM approach, the recall is close to $\rm GENESIS_{MSE}$ at the first time step but drops by a factor of 1.5 to 0.45, at the next time step. Regarding HSS (Fig.~\ref{fig:7}\,f), the GFS forecasts show a time-invariant HSS of around 0.30. Both regression and classification networks have high values of around 0.80 at 30~\si{min}, which are reduced to 0.47 and 0.44 at 240~\si{min}, respectively.

Lastly, we compared the results of GENESIS with a simple random forest (RF) ML model that operates at the pixel level and does not account for the spatiotemporal evolution of the input data. The RF model was configured with a maximum depth of 10 with 20 ensemble decision trees. The results show a significantly lower performance in terms of all the evaluated quality metrics (not shown here). As an example, the GENESIS outperforms RF for $\rm CSI_{1}$ and $\rm CSI_{8}$ by a factor of 1.2 (1.1) and 1.7 (2.5) when MSE (FL) is used as a training loss function.

\subsection{Impacts of Physical Variables}

To quantify the impacts of forecast variables on the performance of the nowcasting machines, we trained identical networks using only sequences of IMERG precipitation (denoted as $\text{GENESIS}^-$) and compared the outputs with those that are informed by the GFS forecasts as well, using both MSE and FL loss functions.

\begin{figure*}[t!]
\includegraphics[width=1\linewidth,]{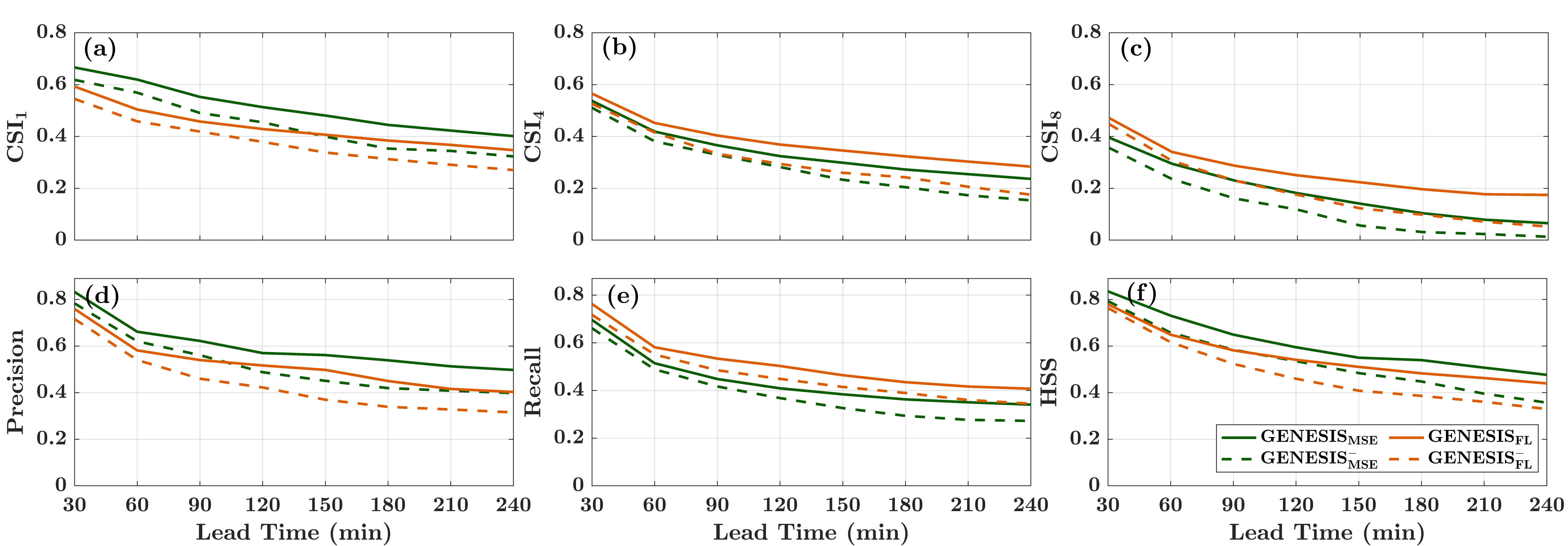}
\caption{Ensemble mean of (a) $\rm CSI_{1}$, (b) $\rm CSI_{4}$, (c) $\rm CSI_{8}$, (d) precision, (e) recall, and (f) HSS for 15,000 independent samples from the test set for $\rm GENESIS_{MSE}$ and $\rm GENESIS_{FL}$ with (solid lines) and without (dashed lines) inclusion of GFS physical variables in the training process.}
\label{fig:8}
\end{figure*}

Figure~\ref{fig:8}\,a shows variation of $\rm CSI_{1}$ over a 4-hour lead time. It can be seen that adding the information content of the GFS physical variables improves the nowcasting skill of the regression (classification) models for $>1$~\si{mm.hr^{-1}} by 8 (10)\% at $T+30$ and by 24 (25)\%  at $T+240$~\si{min}. However, at higher CSI thresholds of 4 and 8 \si{mm.hr^{-1}} (Fig.~\ref{fig:8}\,b--c), the gains become more significant, which is consistent with the results presented in Fig.~\ref{fig:3}, concerning the importance of input features.  

Figure~\ref{fig:8}\,b, shows that for both GENESIS networks at 30~\si{min} lead times, the $\rm CSI_4$ value without and with GFS inputs are approximately close to each other; however, the difference increases to almost 0.1 after 4 hours. The effects of using physical variables become more pronounced in predicting more intense precipitation at longer lead times (Fig.~\ref{fig:8}\,c) as in the regression (classification) network, the $\rm CSI_{8}$ increases from 0.03 to 0.06 (0.02 to 0.11) after 4 hours. Additionally, it is evident that without the inclusion of physical variables, the regression network does not have any skill in predicting high-intensity precipitation after 180~\si{min} while the physically informed network has $\rm CSI_{8}$ value around 0.1 at that lead time. 

Figure~\ref{fig:8}\,d shows that the inclusion of physical variables in the regression (classification) network causes precision to increase by 6 (8)\% at the first and by 22 (29)\% at the last time-step. The improvement in recall (Fig.~\ref{fig:8}\,e) for the regression (classification) network grows by 19 (10)\% when the lead time increases from 30 to 240~\si{min}. In terms of the HSS metric (Fig.~\ref{fig:8}\,f), the same pattern is observed showing that utilizing physical variables can improve the nowcasting skill by around 10\% over longer lead times. 

\subsection{Multiscale Analysis}

\begin{figure*}[t!]
\centering
\includegraphics[width=0.8\linewidth,]{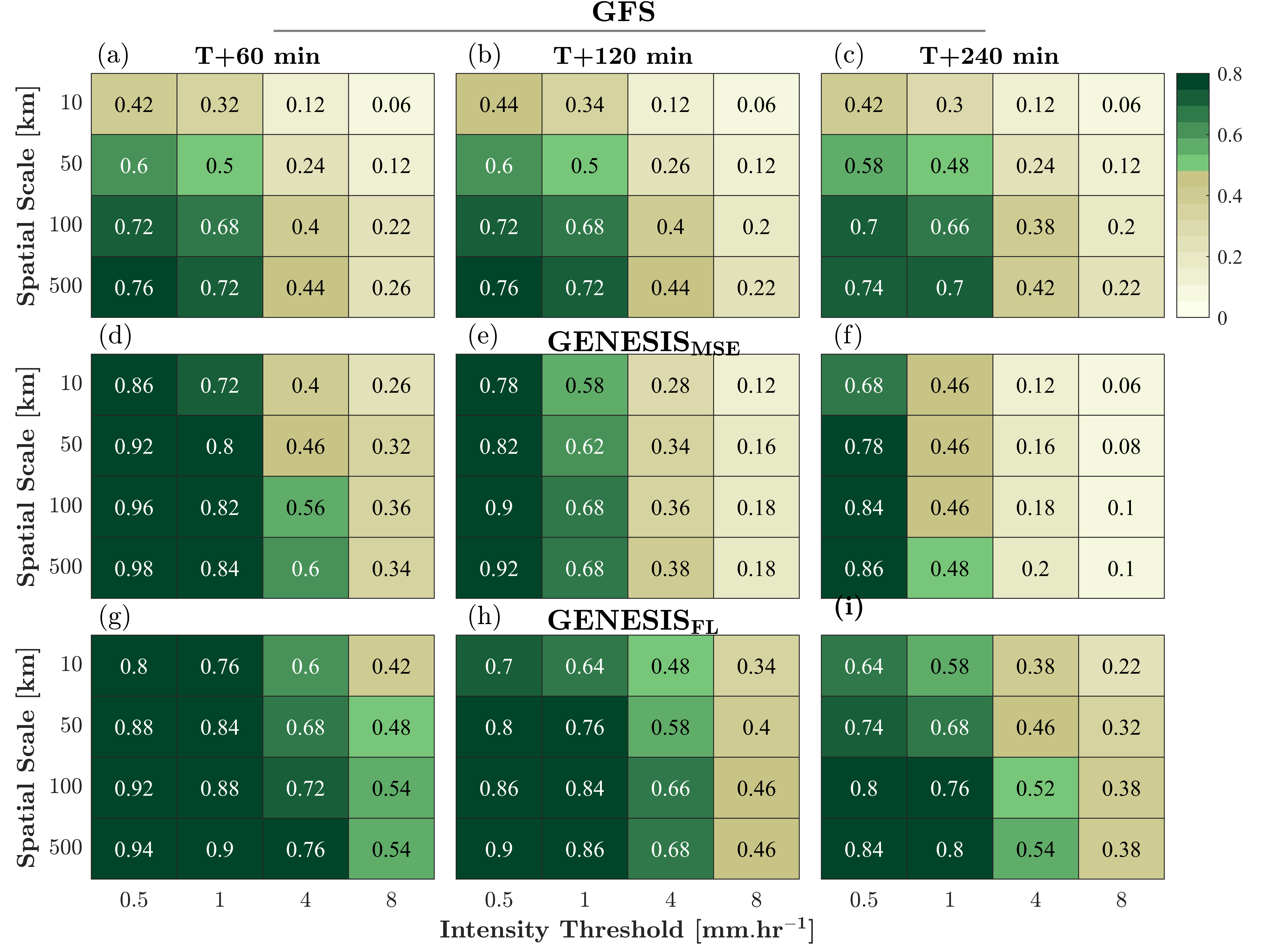}
\caption{Fractions skill scores (FSS) for GFS (a--c), $\rm GENESIS_{MSE}$ (d--f), and $\rm GENESIS_{FL}$ (g--i) at 60, 120, and 240-minute lead times over spatial window sizes of 10, 50, 100, and 500~\si{km} for intensity thresholds of 0.5, 1, 4, and 8~\si{mm.hr^{-1}}.}
\label{fig:9}
\end{figure*}

In this subsection, we aim to use FSS to provide further insights into the dependency of the nowcasting skills on both spatial scales and a precipitation exceedance threshold. As previously explained, this score is obtained by comparing the predicted and reference fractional coverage of precipitation rates, above a certain threshold, within a spatial window defined by a particular scale. The values greater than $0.5+f/2$ are commonly considered skillful, where $f$ is the fraction of the precipitating area over the analyzed domain \citep{skok2016analysis}. Thus if $f$ tends to zero, a value of 0.5 can be considered as a lower bound \citep{mittermaier2010intercomparison}. Figure~\ref{fig:9} shows the mean FSS values for GFS, $\rm GENESIS_{MSE}$, and $\rm GENESIS_{FL}$, obtained over 15,000 independent randomly sampled test events.

As expected, the FSS skills drop with decreasing spatial scales, increasing lead times, and increasing intensity thresholds. As shown in Fig.~\ref{fig:9}\,a--c, once again, the GFS skill is not significantly a function of lead time and remains skillful only for intensity threshold of 0.5 and 1~\si{mm.hr^{-1}} at scales greater than 50~\si{km}. Note that at these intensity thresholds, the main body of the precipitation distribution is dominated by light values which are easier to capture. By increasing the scale from 50 to 500~\si{km}, the skillful FSS values increase roughly from 0.5 to 0.7 across the examined lead times. As expected, the GFS does not show any skill at the native IMERG resolution of 10~\si{km}. 

Figures~\ref{fig:9}\,d--f indicate that the regression network remains skillful at the native IMERG resolution 10~\si{km} when the intensity threshold is 1 (0.5) \si{mm.hr^{-1}} at a 2-hour (4-hour) lead time. However, its performance gradually decreases and becomes closer or even less than GFS for heavier precipitation and longer times. For example, at a scale of 50~\si{km} and rates $>$4~\si{mm.hr^{-1}}, the GFS exhibits a skill of 0.24; however, the skill of the regression network is around 0.16. 

The classification network outperforms the GFS and regression network as the rates increase and the scales decrease (Fig.~\ref{fig:9}\,g--i). We can observe that the FSS ranges from 0.58 to 0.68 for rates greater than 4~\si{mm.hr^{-1}} at a 2-hour lead time, across scales 50--500~\si{km}, while both GFS and the regression network are not skillful. This observation further highlights that the FL-loss function improves the representation of high-intensity precipitation rates at different scales. 

\subsection{Global Error Analysis}

\begin{figure*}[t]
\includegraphics[width=1\linewidth,]{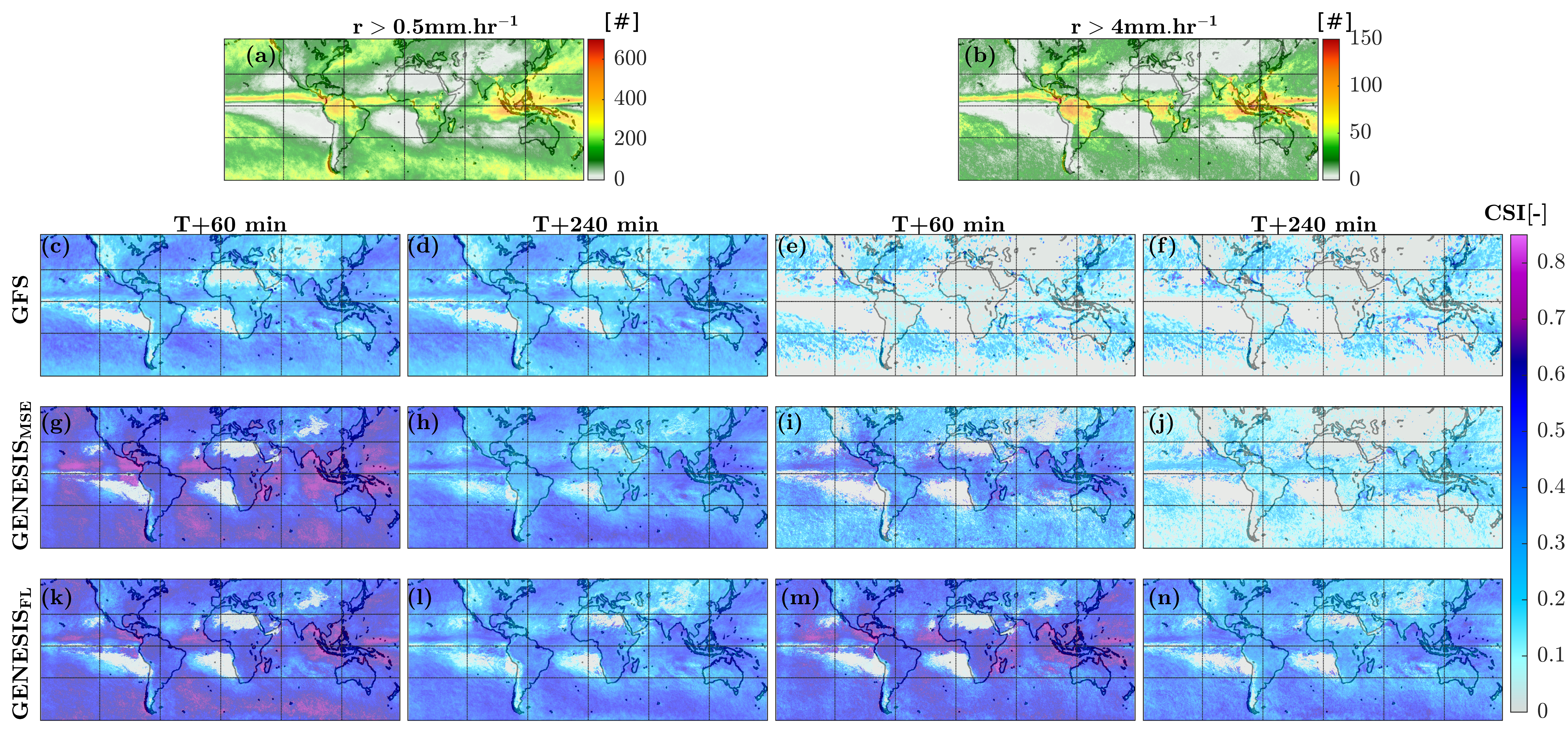}
\caption{Occurrence of precipitation rate above 0.5 and 4 \si{mm.hr^{-1}} (a--b) along with the associated CSI values at 1 and 4 hour lead times for GFS (c--f), $\rm GENESIS_{MSE}$ (g--j), and $\rm GENESIS_{FL}$ (k--n).}
\label{fig:10}
\end{figure*}

This section aims to further evaluate the performance of the nowcasting networks by conducting a quasi-global assessment using the CSI index over one year of the test data, from April 2022 to March 2023. We can provide 48 sets of input tensor blocks (Fig.~\ref{fig:2}) to the nowcasting machines per day. To make the global analysis computationally amenable and reduce the temporal correlation between the samples, we only confine our considerations to six sampled tensor blocks per day with a 4-hour interval. For comparing the results with the reference IMERG, we use the CSI at two different thresholds 0.5 and 4~\si{mm.hr^{-1}} as shown in Fig.~\ref{fig:10}. To ease the discussion, the CSI values are referred to as high (CSI $\geq 0.50$), medium ($0.25 \leq \rm{CSI}<0.50$), and low ($\rm{CSI}<0.25$). 

Figures~\ref{fig:10}\,a--b show the number of precipitation occurrences with rates greater than 0.5 and 4 \si{mm.hr^{-1}}. Precipitation with intensity greater than 0.5~\si{mm.hr^{-1}} follows the general global pattern of precipitation occurrence with the highest rates over the Intertropical Convergence Zone (ITCZ), Amazon, and Congo rainforests, as well as the Maritime Continent, where synoptic low-level convergence is the main lifting mechanism \citep{feng2021global}. Over the mid-latitudes, frequent precipitation mainly occurs over the east coast of North America, East Asia, and the Southern Ocean, where frontal lifting often drives the precipitation \citep{catto2012relating}. As shown, the occurrence of precipitation with intensity greater than 4~\si{mm.hr^{-1}} is more frequent over the tropical regions within $20^\circ$~S-N, where deep convective precipitation is the predominant regime \citep{liu2013climatic}. 

The $\rm CSI_{0.5}$ values for lead times less than 1-hour (Fig.~\ref{fig:10}\,c--d) show that the GFS exhibits low (0.25) and medium (0.35) skills in capturing a large portion of the entire body of the precipitation distribution over land and oceans, respectively. However, by increasing the threshold to 4~\si{mm.hr^{-1}}, the GFS skill in capturing localized and short-lived precipitation over land decreases significantly to almost zero (Fig.~\ref{fig:10}\,e--f). On the other hand, GFS shows slightly improved $\rm CSI_{4}$ over mid-latitudes (0.15) where the extratropical cyclones and their embedded mesoscale frontal sub-systems occur frequently over the Southern Ocean, east coast of North America, and East Asia.

The skill of GFS seems negligible near ITCZ. This may be caused by the misrepresentation of the position of frequent deep and small-scale convective events associated with high-level cumuliform clouds. These clouds are deep and vertically active but small and patchy with a short life cycle of typically less than 2~\si{hr} \citep{rosenfeld1998satellite}. These characteristics make prediction of precipitation events location sensitive \citep{wallace2006atmospheric, zhang2021global}. Analogous to the previous analyses, the results show that the skill is not significantly a function of lead times.

The results in Fig.~\ref{fig:10}\,g--j demonstrate that the regression network has high skills in capturing precipitation rates greater than 0.5~\si{mm.hr^{-1}} both over ocean and land within a 1-hour lead time with mean CSI values of 0.70 and 0.63, respectively. However, the forecast quality declines to medium when the lead time increases by three hours. The regression network, for a 4~\si{mm.hr^{-1}} threshold and a 1-hour lead time, shows high skills over the tropics (0.55) and midum skill over middle latitudes (0.32). Moreover, it can be seen that the skill is higher over the ocean than overland with mean $\rm CSI_4$ values of 0.63 and 0.50, respectively. This pattern may be attributed to a more localized, intense, and transient nature of overland convective precipitation. Increasing the lead time to 4 hours, the regression network shows almost negligible skills over the middle latitudes, where nimbostratus clouds produce relatively long-lived storm events. The skill becomes low (i.e., $\rm CSI_4\approx0.15$) over the tropics, where mostly cumulonimbus clouds produce frequent intense convective precipitation \citep{houze2014cloud}.

Figure~\ref{fig:10}\,k--n indicate that $\rm GENESIS_{FL}$ has approximately similar skill in capturing precipitation rate greater than 0.5~\si{mm.hr^{-1}} compared to $\rm GENESIS_{MSE}$. However, it has high skills for 4~\si{mm.hr^{-1}} threshold at a 1-hour lead time both over the tropics (0.63) and mid-latitudes (0.55). Although, by increasing the lead time to 4 hours, the skill of the classification network decreases to 0.40 and 0.25 over the tropics and mid-latitudes, it still shows a higher skill compared with the regression network in capturing heavier precipitation over both ITCZ and middle latitudes. 

\begin{figure*}[t]
\centering
\includegraphics[width=0.8\linewidth,]{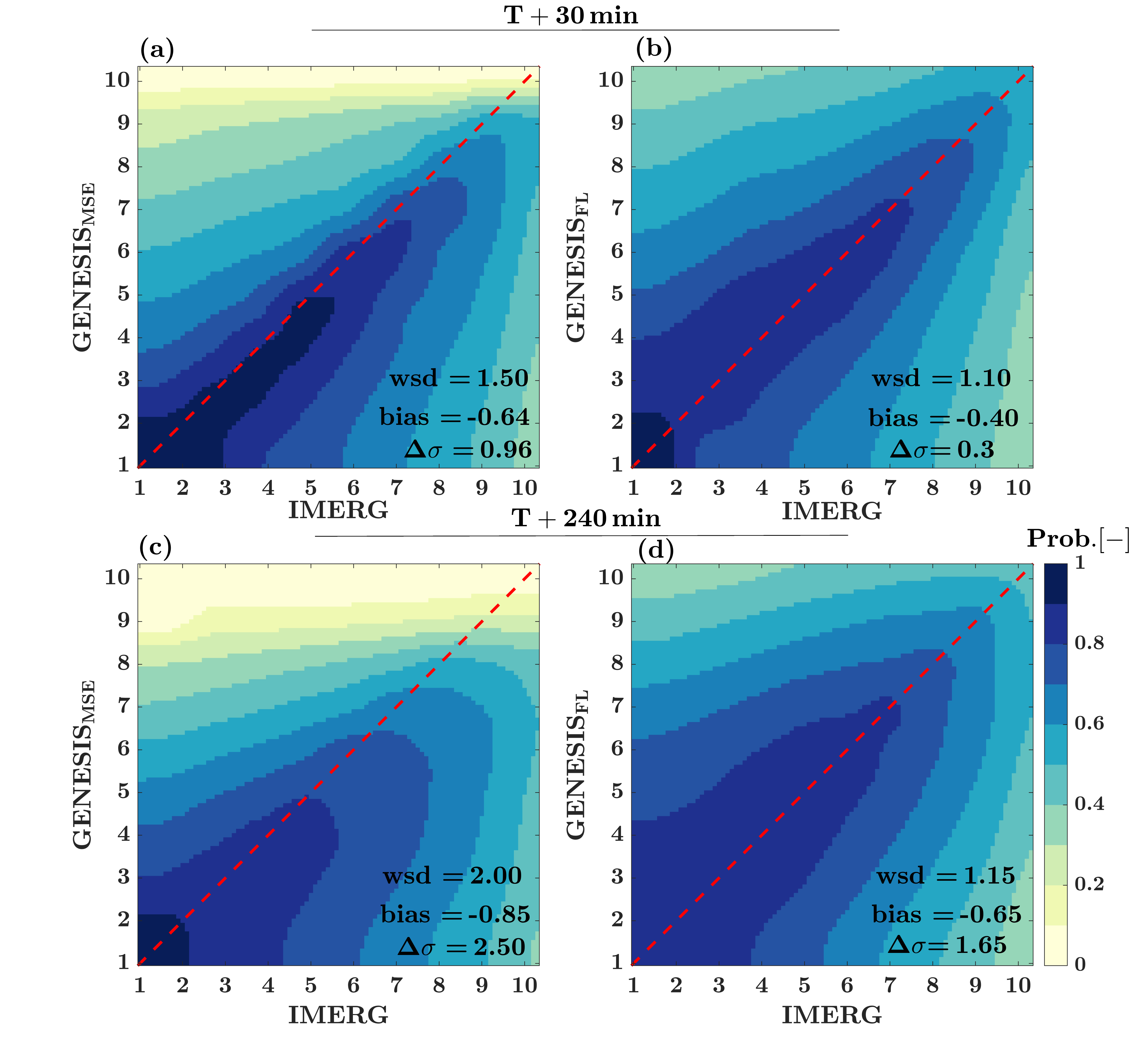}
\caption{Scaled joint precipitation class distribution of $\rm GENESIS_{MSE}$ and $\rm GENESIS_{FL}$ in relation to IMERG at 30 (a--b) and 240~\si{min} (c--d) lead times. Ten classes are defined in a logarithmic scale ranging from 0.1 to 32 \si{mm.hr^{-1}}, assigning rates below (above) 0.1 (32) \si{mm.hr^{-1}} to class 1 (10).}
\label{fig:11}
\end{figure*}

To gain deeper insights into the performance of the nowcasting machines, we examine the joint precipitation class distribution of the nowcasts and IMERG at the first and final lead times. This analysis is conducted using one year of test data, and the results are visualized in Fig.~\ref{fig:11}. To construct this joint histogram, the continuous precipitation rate obtained from IMERG and the regression network are discretized into the defined classes for one-to-one comparisons. To compare the two discrete distributions, three metrics including the conditional bias for precipitation pixels, the difference in standard deviation ($\Delta \sigma$), and the Wasserstein distance (wsd) are reported.

Visual inspection of Fig.~\ref{fig:11}\,a,b shows that the regression network approximates the IMERG data well in the first four classes with a corresponding precipitation rate below 1.6 \si{mm.hr^{-1}} while the classification network has overestimation in those ranges. However, for classes with higher precipitation rates; predictions of the classification network are more consistent with the IMERG. The reported metrics reaffirm the improved performance of the classification compared with the regression network. Overall, the classification network demonstrates less underestimation (i.e., bias = -0.4 \si{mm.hr^{-1}}) and a closer variability to the IMERG, compared to the regression network (i.e., bias = -0.64 \si{mm.hr^{-1}}). Lower values of the Wasserstein distance also indicate that beyond the bias and variability, the classification network is more capable of recovering the entire shape of the precipitation distribution.

Increasing the lead time to 4 hours (Fig.~\ref{fig:11}\,c,d) results in increased dispersion of joint histogram around the diagonal line for both nowcasting networks, which can be attributed to the blurriness effects of the convolution operator at higher lead times. Nonetheless, the classification network continues to outperform the regression network in capturing precipitation at higher rates. As the lead time increases to 4 hours, the Wasserstein distance from the IMERG increases by 33 (5)\% for $\rm GENESIS_{MSE}$ ($\rm GENESIS_{FL}$).

\section{Summary and Conclusion}\label{sec:5}

In this paper, we developed the Global prEcipitation Nowcasting using intEgrated multi-Satellite retrIevalS (GENESIS) through a deep autoencoder neural network by fusing Convolutional LSTM into the U-Net architecture. Dynamic skip connections were proposed to predict the evolution of precipitation fields in time using the recursive generative process ConvLSTM at different encoding blocks. It was demonstrated that the architecture can learn efficiently from the past sequences of IMERG precipitation along with past-to-future sequences of physically related variables to predict precipitation globally every 30 minutes within a 4-hour lead time. The developed network was trained using two different loss functions including mean-squared error (regression) and focal loss (classification) to analyze their effects on capturing extreme precipitation events. The performance of the developed models was assessed against GFS nowcasts -- considering the IMERG as a relative reference.  

The results indicated that recasting the precipitation nowcasting as a classification problem using the FL function can improve capturing the evolution of extreme events beyond a regression loss that relies on the MSE. In particular, on average, the $\rm CSI_{8}$ index increases from 0.18 (regression) to 0.30 (classification) over the entire 4 hours of lead times for the test data set. In contrast, the regression network showed improved detection of rates greater than 1~\si{mm.hr^{-1}}, which are mainly dominated by less intense precipitation. This finding was further substantiated with joint histogram analysis and computing the Wasserstein distance between the precipitation nowcasts and IMERG. In particular, it was shown that the FL loss function better preserves the shape of the precipitation distribution than the mean squared loss function as the lead time grows.

It was unveiled that, on average, the inclusion of the physical variables in the training, including total precipitable water and horizontal near-surface wind velocities, can improve precipitation nowcasting. For example, the recall rate was increased by 12 (10)\% on average over a 4-hour lead time in the regression (classification) network. This improvement was enhanced from 5 (7)\% to 25 (17)\% as the lead time increased from 30 minutes to 4 hours. 

Multiscale analysis of the nowcasting models demonstrated that GFS forecasts were spatially skillful for predicting precipitation rates above 1~\si{mm.hr^{-1}} on spatial scales $s\geq$50~\si{km}, while the nowcasting machines were capable of capturing those events at the native resolution of IMERG at 10~km within the 2-hour lead time. It was shown that the classification network remains skillful for intensities greater than 1 (4)~\si{mm.hr^{-1}} at scales larger than 10 (50)~\si{km} over the entire 4-hour lead times. Also, the spatial error analysis revealed that, compared to the GFS forecasts, the nowcasting machines seem to be effective in extending short-term forecasts of tropical (convective) precipitation more than mid-latitude (frontal) events.

The following future research directions can be envisaged. First, it will be worth investigating how the inclusion of other forecast variables such as cloud liquid and ice water content, as well as surface soil moisture and temperature, can affect the quality of regional precipitation nowcasting. Second, efforts are needed to be devoted to improving the prediction of heavy precipitation at longer lead times perhaps by providing more high-resolution simulation/observational constraints as well as exploring new loss functions that are more adapted to the problem at hand. Third, increasing the number of classes in the classification network to enhance the accuracy of rate range estimation should be considered for future research. Fourth, it will be worth analyzing the performance of training two separate networks for tropical and mid-latitude regions as this may lead to improvements due to contrasting precipitation regimes in these regions. Lastly, to cope with the blurriness effects, one might explore embedding a super-resolution convolutional neural network into the nowcasting machine.

This paper is supported with the corresponding repository on GitHub (\url{https://github.com/reyhaneh-92/GENESIS_Nowcast}), which holds pre-trained GENESIS models and architecture -- developed in the Python 3 programming language using the Keras deep learning library.

\acknowledgments
The first and second authors acknowledge the support from NASA's Remote Sensing Theory program (RST, 80NSSC20K1717), Making Earth System Data Records for Use in Research (MEaSUREs, 80NSSC24M0048) and Global Precipitation Measurement (PMM, 80NSSC22K0596). The third author also acknowledges the support of the NASA Precipitation Measurement Mission program (PMM, 80NSSC22K0594). Moreover, the authors acknowledge the Minnesota Supercomputing Institute (MSI) at the University of Minnesota for providing resources that contributed to the research results reported in this paper (\url{https://www.msi.umn.edu/}).

%
%
\datastatement All the utilized datasets are publicly available. The IMERG and GFS datasets can be downloaded from \url{https://disc.gsfc.nasa.gov/}, and \url{https://noaa-gfs-bdp-pds.s3.amazonaws.com/index.html#gfs.20220101/00/atmos/}, respectively.

\clearpage
\bibliographystyle{ametsocV6}
\bibliography{REF.bib}

\end{document}